\documentclass[11pt]{article}

\usepackage[T1]{fontenc}
\usepackage[utf8]{inputenc}
\usepackage[left=1.05in,right=1.05in,top=1.0in,bottom=1.00in]{geometry}
\usepackage{graphicx}
\usepackage{subcaption}
\usepackage{booktabs}
\usepackage{makecell}
\usepackage{amsmath}
\usepackage{amssymb}
\usepackage{amsfonts}
\usepackage{float}
\usepackage{multirow}
\usepackage{tabularx}
\usepackage{array}
\usepackage{natbib}
\usepackage{microtype}
\usepackage{hyperref}

\hypersetup{hidelinks}

\makeatletter
\renewcommand\subsubsection{\@startsection{subsubsection}{3}{\z@}%
  {-1.5ex \@plus -0.5ex \@minus -.2ex}%
  {0.7ex \@plus .2ex}%
  {\normalfont\normalsize\mdseries\itshape}}
\makeatother

\begin{document}

\thispagestyle{empty}

\begin{center}
{\fontsize{17}{21}\selectfont
\textbf{An Agentic Approach for Active Data Collection, Travel Behavior Modeling, and Weather-Sensitive Demand Prediction}\par}

\vspace{1.4em}

{\large
Narges Ahmadi\textsuperscript{a,*},
Yubo Jiao\textsuperscript{a},
Jônatas Augusto Manzolli\textsuperscript{a},
Jiangbo Yu\textsuperscript{a},
Luis Miranda-Moreno\textsuperscript{a}\par}

\vspace{0.7em}

{\small\itshape
\textsuperscript{a}Department of Civil Engineering, McGill University,
Montreal, Quebec, Canada\par}
\end{center}

\vspace{1.8em}
\hrule
\vspace{0.9em}

\noindent{\large\bfseries Abstract}

\vspace{0.55em}

\noindent

Travel-behavior research increasingly combines digital data collection with predictive modeling, yet these stages are often developed and evaluated separately. This study proposes and applies a three-agent workflow integrating conversational data collection, structured data processing, and behavioral prediction. A chatbot-administered, image-augmented stated-preference survey collected mode choices from 92 student commuters across five predefined weather scenarios, yielding 454 respondent--scenario observations. Weather-related associations were analyzed using a multinomial logit model, while logistic regression and random forest provided machine-learning benchmarks. Nine locally deployed large language models (LLMs), ranging from 2 to 35 billion parameters, were evaluated across four zero-shot prompt-and-context conditions and further extended through persona, few-shot, and vision-based configurations. Cycling was particularly sensitive to adverse weather, while public transit use increased substantially under the Snowy scenario. Random forest achieved 69.6\% five-class accuracy, while the best text-only zero-shot LLM reached 69.9\% without task-specific fitting. Habitual travel information produced the most consistent improvement in LLM prediction, Expert framing generally outperformed Role-Play, and persona information was most beneficial when habitual travel information was unavailable. Few-shot prompting further improved five-class prediction for several models, with gains generally stabilizing after a small number of examples. Vision-capable models were also evaluated using the same weather images presented to respondents, and the best vision-based configuration reached the highest observed five-class accuracy of 71.5\%, indicating that visual context can provide additional predictive information for selected models. Taken together, the findings show that habitual travel information, prompting strategy, demonstrations, and visual context can meaningfully shape LLM-based travel-choice prediction. More broadly, the study demonstrates how conversational survey administration, structured data processing, conventional behavioral modeling, machine learning, and multimodal LLM prediction can be coordinated within a single auditable multi-agent workflow, while broader application requires validation with larger and more representative traveler samples.

\vspace{0.75em}

\noindent\textit{Keywords:}
Large language models, travel behavior, travel survey, conversational agents, mode choice, multi-agent system framework

\vspace{0.85em}
\hrule

\vfill

\noindent\rule{0.36\textwidth}{0.4pt}\\[-0.1em]
{\footnotesize
\textsuperscript{*}Corresponding author.\\
\textit{Email:} \texttt{narges.ahmadi@mail.mcgill.ca}
}

\newpage

\section{Introduction}

Understanding and predicting travel behavior is a fundamental component of transportation planning, encompassing interrelated decisions regarding when, where, and how individuals travel. Among these decisions, mode choice is particularly consequential because it determines how demand is distributed across transportation systems and influences infrastructure needs, environmental impacts, and the effectiveness of policies aimed at encouraging sustainable mobility \citep{mcfadden1974, benakiva1985}. Mode choice is shaped not only by individual and alternative-specific characteristics but also by the context in which travel occurs. Weather is an important example, as precipitation, extreme temperatures, and adverse road conditions can alter the attractiveness of walking and cycling and shift travelers toward transit or private vehicles, with effects that vary across individuals and seasons \citep{hyland2018,bocker2013, saneinejad2012}. Understanding such context-dependent responses therefore requires approaches capable of capturing heterogeneous travel behavior under conditions that may not be adequately represented in observed travel data.

Collecting behavioral data that capture these contextual variations remains challenging. Traditional travel surveys, whether administered through interviews, paper questionnaires, or static web forms, can be costly and time-consuming to implement and scale \citep{krueger2016, bansal2018}. Stated-preference (SP) surveys provide an important alternative by allowing researchers to examine choices under controlled or hypothetical conditions, including weather scenarios that may be difficult to observe systematically in revealed-preference data. However, conventional SP instruments typically describe such situations using fixed textual descriptions, requiring respondents to construct the intended context themselves and limiting the incorporation of richer contextual information \citep{celino2020}. This is particularly relevant for weather-dependent travel choices, for which visual conditions such as precipitation, visibility, snow accumulation, or road conditions may form an important part of the decision environment. These limitations motivate data-collection approaches that can combine scalable survey deployment with richer and more consistent representations of travel contexts.

Recent advances in artificial intelligence (AI) provide new possibilities for addressing these limitations. Conversational chatbots can administer surveys through guided interactions, incorporate respondents' previous answers, and embed multimedia content directly within the survey experience, potentially enabling more flexible and engaging forms of behavioral data collection \citep{yu2025, ren2026, krajcovic2026}. At the same time, generative AI tools can support the construction of predefined visual scenarios, while large language models (LLMs) enable natural-language interaction and the processing of heterogeneous information. Together, these capabilities create opportunities to move beyond static, text-based questionnaires toward conversational and context-rich behavioral surveys. Despite growing applications of conversational AI in survey research, however, chatbot-based data collection remains largely unexplored in travel-behavior and mode-choice research, particularly for stated-preference experiments that integrate image-augmented contextual information.

Parallel advances have occurred in travel-choice modeling. Discrete choice models (DCMs), grounded in random-utility theory, remain central to mode-choice analysis because of their interpretability and ability to estimate behaviorally meaningful relationships \citep{mcfadden1974, benakiva1985, train2009}. Machine-learning (ML) methods have subsequently expanded the predictive toolkit by capturing nonlinear relationships and complex interactions that may be difficult to specify parametrically \citep{hensher2000, ali2026}. More recently, LLMs have introduced a different paradigm in which, rather than estimating a fixed mathematical mapping between explanatory variables and choices, they can interpret traveler characteristics and contextual information expressed through natural-language prompts and generate individual-level predictions with limited or no task-specific training \citep{zhao2023, brown2020}. This flexibility also makes it possible to vary the information provided to the model; for example, demographic characteristics, travel history, examples of previous decisions, or contextual descriptions, through alternative prompting configurations. Yet two important questions remain insufficiently examined in travel-behavior prediction. The first concerns how prediction performance changes across systematically designed LLM configurations and prompting strategies. The second is whether vision-capable LLMs can directly incorporate visual representations of travel conditions, such as the same weather images presented to human respondents, rather than relying exclusively on textual descriptions.

Beyond prediction, LLMs have increasingly been studied as \emph{synthetic respondents}, serving as computational representations of individuals that generate decisions conditioned on specified personas and choice contexts. This line of research initially developed in economics, political science, and survey research, where LLM-generated responses have been compared with human decisions \citep{horton2023, argyle2023}. Recent transportation studies similarly suggest that conditioning LLMs on traveler characteristics, behavioral histories, or latent attitudes can improve their correspondence with observed travel choices \citep{liu2025framework, sameen2025}. A related development is the emergence of multi-agent systems in which specialized or persona-based LLM agents perform complementary functions or represent heterogeneous decision-makers at scale \citep{liu2025framework, liu2026dual, chu2025}. Nevertheless, LLM-generated behavior does not necessarily reproduce human decision-making reliably, particularly under limited-context or zero-shot settings \citep{liu2024ssrn,bisbee2024, hullman2026}. Despite these developments across other behavioral domains, the integration of specialized agents for data collection, data processing, and LLM-based behavioral prediction within a common \emph{multi-agent workflow} remains largely unexplored for travel behavior and individual mode-choice modeling.

This study addresses these gaps through a multi-agent workflow that connects AI-assisted behavioral data collection with travel-choice modeling and LLM-based prediction. The paper makes four main contributions. First, it develops a conversational, image-augmented SP survey in which an AI-based chatbot collects traveler characteristics and repeated mode choices across predefined weather scenarios, providing a structured dataset specifically designed for downstream behavioral prediction. Second, it provides empirical evidence on weather-related variation in commuter mode choice by examining how the same individuals change their stated choices across systematically varied weather scenarios. Third, it systematically evaluates LLMs as mode-choice predictors under multiple prediction configurations, including zero-shot, few-shot, persona- and travel-history-enhanced prompting, as well as vision-based prediction using the same weather images presented to respondents, and benchmarks their performance against conventional DCM and ML approaches. Fourth, it demonstrates a \emph{multi-agent workflow} that links conversational survey administration, structured data processing, and parallel discrete-choice, machine-learning, and LLM-based modeling within an integrated data-collection-to-prediction process, providing a reproducible basis for exploring AI-assisted travel-behavior research.

The remainder of the paper is organized as follows. Section~2 reviews related work on travel-behavior data collection and conversational agents, mode-choice modeling, LLM-based choice prediction, and multi-agent applications. Section~3 presents the proposed multi-agent methodology. Section~4 describes the case-study implementation, and Section~5 reports the descriptive and behavioral-modeling results and the LLM prediction experiments. The subsequent sections discuss the implications and limitations of the findings and conclude with directions for future research.

\section{Background and Related Work}

\subsection{Travel Behavior and Mode Choice}

Mode choice is one of the oldest and most heavily modeled problems in transportation research, and the multinomial logit (MNL) model has been its dominant framework for decades \citep{mcfadden1974, benakiva1985}. Grounded in random utility theory, MNL and its extensions such as nested and mixed logit remain the standard tool for estimating how travel time, cost, and individual characteristics translate into a discrete travel choice, in large part because their coefficients admit a direct behavioral interpretation \citep{train2009}. Machine learning (ML) classifiers, including random forests, gradient boosting, and support vector machines, have become an increasingly common complement to discrete choice models. Comparative studies have repeatedly found that ML classifiers can match or exceed MNL on predictive accuracy, particularly when the feature set is large or the choice set is imbalanced, though the resulting models are harder to interpret in behavioral terms \citep{hensher2000, ali2026}.

Both model families face a common challenge: traveler preferences are not fixed but vary with weather, season, the built environment, and individual heterogeneity. Weather in particular has a well-documented association with active-mode choice, with cold, wet, and low-visibility conditions suppressing walking and cycling and shifting travelers toward transit or driving \citep{bocker2013, saneinejad2012, hyland2018}. Capturing this sensitivity often requires stated-preference (SP) methods that vary conditions directly, because revealed-preference data may not span the full range of weather a respondent might encounter \citep{bansal2018, krueger2016}. Both MNL and ML models, however, require labeled data, and their predictive performance may deteriorate when application conditions differ from those represented during model development. This limitation motivates the growing interest, reviewed in Section 2.3, in LLM-based choice prediction.

\subsection{Data Collection Methods in Travel Behavior Research}

Travel behavior data has traditionally been collected through paper diaries, telephone interviews, and static web-based surveys. Each instrument trades off cost, reach, and depth; for example, paper surveys are labor-intensive to process, telephone interviews are expensive to scale, and static web forms, while cheap to distribute, present every respondent with the same fixed sequence of questions regardless of their answers \citep{bonnel2018}. Conversational and digital instruments relax some of these constraints. Chatbot-administered surveys can branch dynamically on prior responses, and early evaluations report that respondents find conversational interfaces more engaging than static forms without a loss in data quality \citep{celino2020}. Recent transportation-specific work has begun to apply modular AI agents to survey administration and interviewing, arguing that conversational agents can improve engagement, transparency, and cost efficiency relative to conventional instruments \citep{yu2025}, and adaptive, reinforcement-learning-driven conversational survey designs have been proposed to further personalize the interview flow \citep{tang2025}.

Stated-preference tasks have also begun to incorporate richer visual aids, from virtual reality environments to video and photorealistic imagery, on the premise that a visual scenario elicits a more ecologically valid response than a text description alone \citep{farooq2024}. Pilot studies using video-based conversational chatbots to elicit perceived cycling safety \citep{ren2026} and photorealistic embodied conversational agents in survey research \citep{krajcovic2026} point in the same direction, but remain limited in scale. Despite this progress, most chatbot instruments in transportation are still used for information delivery, such as trip planning or transit information, rather than for behavioral data collection, and few stated-preference studies embed rich media such as generated images directly into a chatbot-administered choice task. This is the gap the data collection agent in Section 4.1 is designed to address.

\subsection{LLMs for Behavioral Simulation}

Large language models have recently been explored as tools for behavioral simulation across several fields. In economics, they have been proposed as simulated agents \citep{horton2023}; in political science, they have been used to reproduce survey samples \citep{argyle2023} and support population-scale agent-based simulations \citep{park2024generative}; and in market research and consumer behavior, they have been applied to segmentation and behavioral simulation \citep{li2025segmentation, brand2023, estevez2025}. Related studies have examined trust behavior \citep{jia2024} and large-scale replication of psychology and management experiments \citep{cui2025}, illustrating the broader potential of LLMs as synthetic participants. At the same time, this literature also raises important concerns. Replication studies show that LLM-generated survey responses may diverge from human data in ways that are difficult to detect \citep{bisbee2024}, while methodological critiques caution against treating simulated responses as equivalent to human-subject evidence without independent validation \citep{hullman2026}. Research on persona-based and character-consistent prompting similarly suggests that adopting a simulated identity may either improve or distort behavioral responses \citep{chen2024oscars, xu2024character}.

Within transportation, a small but growing body of work investigates whether LLMs can predict individual travel choices without being trained on the target dataset. Applications include mode choice prediction \citep{mo2023}, alternative-set evaluation \citep{nishida2025}, and behaviorally informed ridesourcing choice modeling \citep{sameen2025}. Other studies examine whether persona-based representations improve agreement with observed travel choices \citep{liu2025persona}, whether LLMs can recover travelers' valuation of time \citep{yan2026}, and whether they can reproduce stated airline passenger preferences \citep{voltesdorta2025}. An early working paper further suggests that LLMs can capture a meaningful share of the predictive signal in observed mode choice data without dataset-specific training \citep{liu2024ssrn}. Despite these advances, existing studies generally examine individual prompting or modeling choices in isolation. A systematic comparison of context richness, model scale, persona information, few-shot learning, and visual inputs within a single travel-behavior prediction task remains limited. In particular, the use of the same visual travel conditions presented to human respondents as direct inputs to vision-capable LLMs remains unexplored, constituting an additional gap addressed by this study.

\subsection{Multi-Agent Systems and LLM-Agent Frameworks}

Outside transportation, LLM-based multi-agent systems have been proposed for tasks ranging from simulating marketing and consumer behavior \citep{chu2025} to sandboxed economic modeling of consumer preference alignment \citep{wu2026} and constructing virtual organizational decision-makers for management research \citep{garzonvico2026}. These systems typically decompose a task among specialized agents that exchange intermediate results, drawing on the broader software-engineering literature on tool use and orchestration in autonomous LLM agents \citep{zhao2023}. Within transportation, this framing remains largely conceptual. Recent work has proposed frameworks for LLM-agent-based modeling of transportation systems \citep{liu2025framework} and dual-agent approaches for aligning LLM agents with human learning and adjustment behavior \citep{liu2026dual}. Few studies, however, have implemented and evaluated an end-to-end workflow from survey design through data processing to behavioral prediction on a common empirical dataset. The present paper addresses this gap by instantiating the three stages as specialized agents and applying them to the same commuter mode-choice dataset.

\section{Methodology}
\label{sec:methodology}

\subsection{Framework Overview}
\label{sec:framework_overview}
\begin{figure}[t]
\centering
\includegraphics[width=0.98\linewidth]{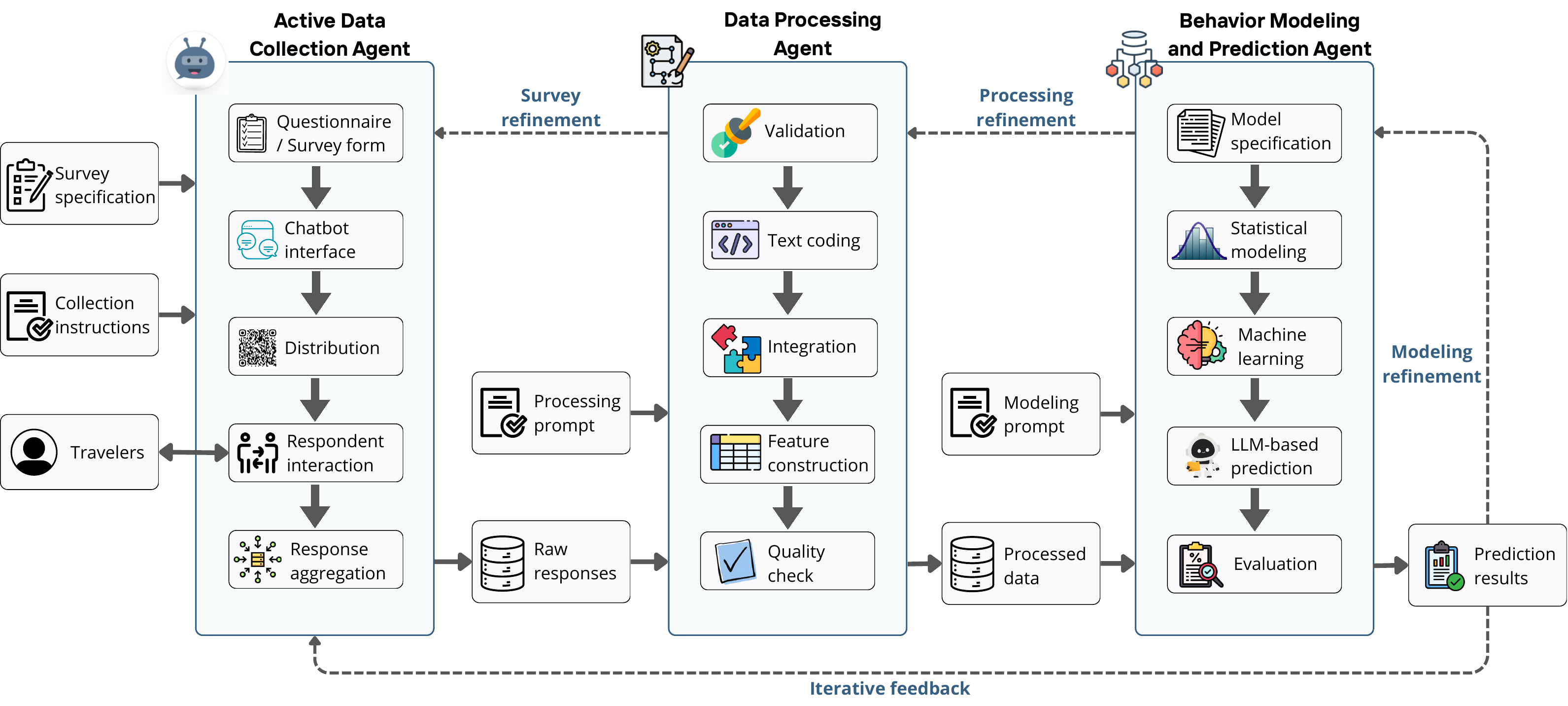}
\caption{Multi-agent framework for commuter travel-behavior research. Solid arrows denote forward data and artifact flows, whereas dashed arrows denote refinements implemented in a subsequent workflow iteration.}
\label{fig:framework}
\end{figure}

We propose a multi-agent framework that integrates data collection, data processing, and behavioral modeling for commuter mode-choice research. The framework consists of three specialized agents: a Data Collection Agent, a Data Processing Agent, and a Data Modeling Agent. The agents perform distinct research functions and exchange information through standardized data objects and explicit input--output interfaces. Data transfer, version control, and execution records are organized through a shared workspace and predefined workflow rules. These supporting functions are system infrastructure rather than additional agents. Figure~\ref{fig:framework} illustrates the forward workflow and the associated refinement paths.

In this study, an agent is defined as a goal-directed and researcher-supervised software component that can interpret user instructions, maintain task state, select from an authorized set of tools, and produce structured and auditable outputs. An agent is therefore not restricted to an LLM. Depending on the task, it may combine an LLM with deterministic rules, database operations, statistical software, machine-learning algorithms, or general-purpose programming tools. The proposed framework is consequently a multi-agent research workflow rather than an LLM-only pipeline. The term \emph{agent} in this framework refers to a software agent that supports the research process and should not be confused with a traveler agent in conventional agent-based transport simulation. Human respondents remain the source of stated survey choices, while researchers retain control over the survey constructs, data-processing requirements, model specifications, and interpretation of results. Throughout this paper, \emph{active data collection} refers to stateful conversational survey administration with predefined interaction logic and structured response capture; it does not refer to active learning or adaptive sampling. Weather-sensitive demand prediction refers to individual mode-choice prediction across the five predefined weather scenarios.

The framework follows the information flow:

\begin{equation}
\mathcal{Q}
\xrightarrow{
\mathcal{A}_{\mathrm{col}}
\left(
\cdot;
\mathbf{u}_{\mathrm{col}}
\right)
}
\left(
\mathcal{D}^{\mathrm{raw}},
\mathcal{B}_{\mathrm{col}}
\right)
\xrightarrow{
\mathcal{A}_{\mathrm{pro}}
\left(
\cdot,\mathcal{E};
\mathbf{u}_{\mathrm{pro}}
\right)
}
\left(
\mathcal{D}^{\mathrm{pro}},
\mathcal{Z}_{\mathrm{pro}}
\right)
\xrightarrow{
\mathcal{A}_{\mathrm{mod}}
\left(
\cdot;
\mathbf{u}_{\mathrm{mod}}
\right)
}
\left(
\widehat{\mathcal{M}},
\mathcal{R}_{\mathrm{mod}}
\right)
\label{eq:overall_workflow}
\end{equation}

\noindent where $\mathcal{Q}$ denotes the researcher-defined survey specification; $\mathbf{u}_{\mathrm{col}}$, $\mathbf{u}_{\mathrm{pro}}$, and $\mathbf{u}_{\mathrm{mod}}$ denote the user instructions provided to the three agents; $\mathcal{D}^{\mathrm{raw}}$ is the collected raw dataset; $\mathcal{B}_{\mathrm{col}}$ is the survey deployment package; $\mathcal{E}$ represents optional external data; $\mathcal{D}^{\mathrm{pro}}$ is the processed dataset; $\mathcal{Z}_{\mathrm{pro}}$ contains supplementary processing outputs; $\widehat{\mathcal{M}}$ is the fitted model or set of models; and $\mathcal{R}_{\mathrm{mod}}$ contains model estimates, predictions, diagnostics, and evaluation results.

Equation~\eqref{eq:overall_workflow} represents one forward execution pass. The dashed arrows in Figure~\ref{fig:framework} indicate diagnostic feedback that may initiate a revised survey, processing, or modeling specification. Such revisions do not automatically modify an active study; they are reviewed by the researchers and implemented as a new workflow version. The framework treats raw responses as immutable and reserves held-out test outcomes for evaluation rather than revision of the configuration assessed on that test set.

This separation assigns a distinct responsibility to each agent. The Data Collection Agent converts a survey specification into a deployable conversational interface and collects responses. The Data Processing Agent converts raw records into analysis-ready data. The Data Modeling Agent translates the research objective into an executable modeling workflow and returns fitted models and associated results. The decomposition reduces the need for a single general-purpose agent to perform all tasks and makes errors easier to identify at their source.

\begin{table}[t]
\centering
\caption{Roles, interfaces, and supported functions of the three agents.}
\label{tab:agent_roles}
\footnotesize
\renewcommand{\arraystretch}{1.15}
\begin{tabularx}{\textwidth}{
    >{\raggedright\arraybackslash}p{0.18\textwidth}
    >{\raggedright\arraybackslash}p{0.22\textwidth}
    >{\raggedright\arraybackslash}p{0.25\textwidth}
    >{\raggedright\arraybackslash}X}
\toprule
Agent & Main inputs & Authorized functions & Main outputs \\
\midrule
Data Collection Agent
&
Survey specification and collection instructions
&
Chatbot construction, survey logic implementation, procedural question answering, response validation, clarification, deployment, and response aggregation
&
Deployable survey interface, distribution link or QR code, raw responses, interaction logs, and collection metadata
\\

Data Processing Agent
&
Raw survey data, optional external data, codebook, and processing instructions
&
De-identification, validation, recoding, missing-data handling, text coding, data integration, feature construction, and quality assessment
&
Processed dataset, quality flags, derived variables, supplementary data, and processing records
\\

Data Modeling Agent
&
Processed data, modeling instructions, candidate model library, and evaluation requirements
&
Model specification, code generation, statistical estimation, machine-learning training, LLM-based choice prediction, validation, and result organization
&
Fitted model, parameter estimates, predictions, uncertainty measures, diagnostics, evaluation metrics, and structured results
\\
\bottomrule
\end{tabularx}
\vspace{2pt}
\parbox{\textwidth}{\footnotesize\textit{Note:} The table summarizes functions supported by the general framework. The subset implemented in the present case study is described in Section~4.}
\end{table}

\subsection{General Agent Representation and Workflow Orchestration}
\label{sec:agent_representation}

The following formulation describes functions supported by the general framework. Section~4 identifies the subset instantiated in the present case study; not every supported function was exercised or empirically evaluated in the current application.

Let $k\in\{\mathrm{col},\mathrm{pro},\mathrm{mod}\}$ index the three agents. Each agent is represented as a mapping:

\begin{equation}
\mathcal{A}_{k}:
\left(
\mathbf{x}_{k},
\mathbf{u}_{k},
\mathbf{s}_{k};
\boldsymbol{\Theta}_{k}
\right)
\mapsto
\left(
\mathbf{y}_{k},
\mathbf{s}'_{k},
\boldsymbol{\ell}_{k}
\right)
\label{eq:generic_agent}
\end{equation}

\noindent where $\mathbf{x}_{k}$ denotes task-specific input artifacts, $\mathbf{u}_{k}$ denotes the user instruction, $\mathbf{s}_{k}$ is the current task state, and $\boldsymbol{\Theta}_{k}$ contains the agent configuration, such as its prompt templates, model settings, tool permissions, and termination rules. The outputs include the task result $\mathbf{y}_{k}$, the updated state $\mathbf{s}'_{k}$, and an execution log $\boldsymbol{\ell}_{k}$.

At execution step $t$, the agent selects a tool or operation according to:

\begin{equation}
\tau_{k,t}
=
\pi_{k}
\left(
\mathbf{s}_{k,t},
\mathbf{x}_{k,t},
\mathbf{u}_{k};
\boldsymbol{\Theta}_{k}
\right),
\qquad
\tau_{k,t}\in\mathbb{T}_{k},
\label{eq:tool_selection}
\end{equation}

\noindent where $\pi_{k}$ is the agent policy and $\mathbb{T}_{k}$ is its authorized tool library. The selected tool produces an intermediate output and updates the task state:

\begin{equation}
\left(
\mathbf{o}_{k,t},
\mathbf{s}_{k,t+1}
\right)
=
g_{\tau_{k,t}}
\left(
\mathbf{x}_{k,t},
\mathbf{s}_{k,t}
\right)
\label{eq:state_update}
\end{equation}

This formulation allows an agent to use different tools for different subtasks. For example, the Data Processing Agent may use deterministic Python code for range checks, an LLM for coding open-ended responses, and a database operation for joining external data. Tool selection is constrained by the agent's permissions and by the output schema required by the next stage.

The workflow orchestrator initiates each agent, verifies whether its required inputs are available, and passes only validated artifacts to the next stage. Direct, unrestricted natural-language communication between agents is avoided. Instead, the agents communicate through typed data objects, including datasets, codebooks, configuration files, model specifications, and execution records. This design improves interoperability and limits the propagation of unsupported agent outputs.

\subsection{Data Collection Agent}
\label{sec:data_collection_agent}

The Data Collection Agent converts a researcher-defined questionnaire into a deployable conversational survey, manages survey distribution, collects respondent inputs, and aggregates the resulting records. Its design builds on modular conversational survey systems in which the user interface, prompts, knowledge resources, session variables, and conversational logic are explicitly configured.

\subsubsection{Survey specification}

The survey remains a researcher-defined measurement instrument. Let $\mathcal{Q} = \left\{ q_{\ell} \right\}_{\ell=1}^{L}$ denote a survey containing $L$ question objects. Each question is represented as:

\begin{equation}
q_{\ell}
=
\left(
c_{\ell},
w_{\ell},
r_{\ell},
\mathcal{O}_{\ell},
b_{\ell},
v_{\ell}
\right)
\label{eq:question_schema}
\end{equation}

\noindent where $c_{\ell}$ is the behavioral construct, $w_{\ell}$ is the question wording, $r_{\ell}$ is the response type, $\mathcal{O}_{\ell}$ is the set of allowable options when applicable, $b_{\ell}$ defines branching and display conditions, and $v_{\ell}$ defines validation rules. Response types may include single-choice, multiple-choice, numerical input, Likert-scale, text, voice, or image-assisted questions.

The deployable chatbot is generated as:
\begin{equation}
\mathcal{C}
=
\Gamma_{\mathrm{col}}
\left(
\mathcal{Q},
\mathbf{u}_{\mathrm{col}},
\boldsymbol{\Omega}_{\mathrm{col}}
\right)
\label{eq:chatbot_generation}
\end{equation}

\noindent where $\Gamma_{\mathrm{col}}$ is the chatbot-construction procedure and $\boldsymbol{\Omega}_{\mathrm{col}}$ contains the available interface components, prompt templates, knowledge resources, image assets, and deployment tools. The output $\mathcal{C}$ contains the survey interface, question sequence, branching logic, data-storage schema, and interaction rules.

The agent may use an LLM to generate interface code, reformulate researcher instructions into chatbot logic, or produce an initial implementation for review. However, constructs, response scales, experimental attributes, and mandatory questions cannot be changed without explicit researcher approval. This restriction preserves measurement consistency across respondents.

\subsubsection{Conversational interaction and response validation}

For respondent $i$ and question $\ell$, let $h_{i\ell}^{(r)}$ denote the interaction history after clarification round $r$. The agent evaluates whether the accumulated response satisfies the question requirement:

\begin{equation}
\xi_{i\ell}^{(r)}
=
V_{\mathrm{col}}
\left(
q_{\ell},
h_{i\ell}^{(r)};
\boldsymbol{\Theta}_{\mathrm{col}}
\right),
\qquad
\xi_{i\ell}^{(r)}\in\{0,1\}
\label{eq:response_sufficiency}
\end{equation}

\noindent where $\xi_{i\ell}^{(r)}=1$ indicates that the response is sufficiently complete and valid. If $\xi_{i\ell}^{(r)}=0$ and the maximum number of clarification rounds has not been reached, the agent generates a question-specific clarification:

\begin{equation}
q_{i\ell}^{c,(r+1)}
=
F_{\mathrm{col}}
\left(
q_{\ell},
h_{i\ell}^{(r)},
\mathbf{u}_{\mathrm{col}}
\right)
\label{eq:clarification_question}
\end{equation}

The process continues until the response is accepted or the predefined clarification limit $R_{\max}$ is reached. The final structured answer is:

\begin{equation}
x_{i\ell}^{\mathrm{col}}
=
E_{\mathrm{col}}
\left(
h_{i\ell}^{(R_{i\ell})},
q_{\ell}
\right),
\qquad
R_{i\ell}\leq R_{\max}
\label{eq:response_extraction}
\end{equation}

\noindent where $E_{\mathrm{col}}$ maps the interaction history to the survey schema. For fixed-choice questions, deterministic option matching is used whenever possible. LLM-based mapping is used only when an answer is provided in unrestricted text or speech and cannot be mapped using deterministic rules.

The agent may also provide procedural assistance, such as explaining a survey term, repeating a question, changing the interaction language, or allowing a respondent to revise an earlier answer. Such assistance must not recommend a travel mode, signal a preferred response, or otherwise alter the intended behavioral task.

\subsubsection{Deployment and raw-data output}

After the survey has been approved, the agent generates a deployment bundle:

\begin{equation}
\mathcal{B}_{\mathrm{col}}
=
\left(
\mathrm{URL},
\mathrm{QR},
\mathrm{version},
\mathrm{recruitment\ material}
\right),
\label{eq:deployment_bundle}
\end{equation}

\noindent which may contain a public or tokenized survey link, a QR code, survey version information, and standardized recruitment materials.

For respondent $i$, the collection record is represented as:

\begin{equation}
d_{i}^{\mathrm{raw}}
=
\left(
\mathbf{x}_{i}^{\mathrm{col}},
\mathbf{h}_{i},
\mathbf{m}_{i},
\mathbf{p}_{i}
\right),
\label{eq:raw_respondent_record}
\end{equation}

\noindent where $\mathbf{x}_{i}^{\mathrm{col}}$ contains the extracted answers, $\mathbf{h}_{i}$ contains the raw interaction history, $\mathbf{m}_{i}$ contains response and session metadata, and $\mathbf{p}_{i}$ contains provenance information such as the survey version, prompt version, interface version, and timestamps. The complete raw dataset is:

\begin{equation}
\mathcal{D}^{\mathrm{raw}}
=
\bigcup_{i=1}^{N}
d_{i}^{\mathrm{raw}}.
\label{eq:raw_dataset}
\end{equation}

\noindent The raw dataset is treated as immutable. Later corrections, exclusions, or recoding decisions are stored as separate processing operations rather than overwriting the original responses.

\subsection{Data Processing Agent}
\label{sec:data_processing_agent}

The Data Processing Agent converts the raw records into an analysis-ready dataset based on the user instruction $\mathbf{u}_{\mathrm{pro}}$. Its functions may include de-identification, schema validation, data-type conversion, missing-data handling, text coding, consistency checking, feature construction, and integration with external datasets. It may also produce supplementary outputs such as a codebook, data-quality report, descriptive summaries, or a list of unresolved records.

Let $\mathcal{E}$ denote optional external data, such as weather records, spatial attributes, transit service information, or zonal characteristics. The initial processing object is:

\begin{equation}
\mathcal{D}^{(0)}
=
J
\left(
\mathcal{D}^{\mathrm{raw}},
\mathcal{E};
\mathcal{K}_{\mathrm{join}}
\right)
\label{eq:data_initialization}
\end{equation}

\noindent where $J(\cdot)$ is a controlled integration operator and $\mathcal{K}_{\mathrm{join}}$ specifies the permitted identifiers, temporal alignment rules, spatial matching rules, and source metadata. External data are not merged unless their source and matching procedure can be recorded.

The processing agent constructs an ordered sequence of operations:

\begin{equation}
\mathcal{P}
=
\left(
T_{1},
T_{2},
\ldots,
T_{H}
\right),
\qquad
T_{h}\in\mathbb{T}_{\mathrm{pro}},
\label{eq:processing_plan}
\end{equation}

\noindent where $\mathbb{T}_{\mathrm{pro}}$ is the authorized processing-tool library. At step $h$, the agent selects an operation according to:

\begin{equation}
z_{h}
=
\pi_{\mathrm{pro}}
\left(
\mathbf{u}_{\mathrm{pro}},
\mathbf{s}_{h},
\nu\left(\mathcal{D}^{(h-1)}\right)
\right)
\label{eq:processing_tool_selection}
\end{equation}

\noindent where $\nu(\cdot)$ returns the current dataset profile, including variable types, missingness, range violations, duplicate records, and unresolved text fields. The dataset is then updated as:

\begin{equation}
\mathcal{D}^{(h)}
=
T_{z_h}
\left(
\mathcal{D}^{(h-1)};
\boldsymbol{\theta}_{h}
\right),
\qquad
h=1,\ldots,H
\label{eq:processing_update}
\end{equation}

\noindent where $\boldsymbol{\theta}_{h}$ contains the operation-specific parameters. The final processed dataset is:

\begin{equation}
\mathcal{D}^{\mathrm{pro}}
=
\mathcal{D}^{(H)}
\label{eq:processed_dataset}
\end{equation}

\noindent Deterministic operations are preferred when the processing rule can be stated explicitly. Examples include numerical range checks, category recoding, duplicate removal, unit conversion, date processing, and table joins. LLMs are used for language-dependent tasks for which fixed rules are insufficient, such as mapping an open-ended explanation to a predefined coding scheme or extracting a stated constraint from a conversation.

For an unstructured response $r_{i\ell}$ and an admissible code set $\mathcal{C}_{\ell}$, an LLM-assisted coding operation returns:

\begin{equation}
\left(
\widetilde{x}_{i\ell},
\gamma_{i\ell}
\right)
=
H_{\psi}
\left(
r_{i\ell},
\mathcal{C}_{\ell},
\mathbf{u}_{\mathrm{pro}}
\right)
\label{eq:llm_coding}
\end{equation}

\noindent where $\widetilde{x}_{i\ell}$ is the proposed code and $\gamma_{i\ell}$ is a task-specific confidence or validation score. The stored value is determined by:

\begin{equation}
x_{i\ell}^{\mathrm{pro}}
=
\begin{cases}
g_{\ell}(r_{i\ell}),
&
\text{if deterministic parsing succeeds},
\\[4pt]
\widetilde{x}_{i\ell},
&
\text{if }
\widetilde{x}_{i\ell}\in\mathcal{C}_{\ell}
\text{ and }
\gamma_{i\ell}\geq\delta_{\ell},
\\[4pt]
\mathrm{NA},
&
\text{otherwise}
\end{cases}
\label{eq:hybrid_coding_rule}
\end{equation}

\noindent where $g_{\ell}$ is the deterministic coding rule and $\delta_{\ell}$ is the acceptance threshold. Records that do not satisfy the rule are flagged for human review instead of being assigned an unverified value.

The processed dataset may be represented at the respondent--scenario level. For respondent $i$ and commuting scenario $s$, define:

\begin{equation}
d_{is}^{\mathrm{pro}}
=
\left(
\mathbf{p}_{i},
\mathbf{z}_{is},
\mathbf{a}_{is},
y_{is},
\mathbf{q}_{is},
\mathbf{v}_{is}
\right)
\label{eq:respondent_scenario_record}
\end{equation}

\noindent where $\mathbf{p}_{i}$ contains respondent characteristics, $\mathbf{z}_{is}$ contains scenario and alternative attributes, $\mathbf{a}_{is}$ contains alternative-availability indicators, $y_{is}$ is the observed mode choice, $\mathbf{q}_{is}$ contains data-quality flags, and $\mathbf{v}_{is}$ contains data provenance. The resulting dataset is:

\begin{equation}
\mathcal{D}^{\mathrm{pro}}
=
\left\{
d_{is}^{\mathrm{pro}}
:
i=1,\ldots,N,\;
s=1,\ldots,S_i
\right\}
\label{eq:respondent_scenario_dataset}
\end{equation}

In addition to $\mathcal{D}^{\mathrm{pro}}$, the agent returns:

\begin{equation}
\mathcal{Z}_{\mathrm{pro}}
=
\left(
\mathcal{C}_{\mathrm{book}},
\mathcal{F}_{\mathrm{quality}},
\mathcal{S}_{\mathrm{summary}},
\mathcal{U}_{\mathrm{unresolved}}
\right)
\label{eq:processing_outputs}
\end{equation}

\noindent where $\mathcal{C}_{\mathrm{book}}$ is the final codebook, $\mathcal{F}_{\mathrm{quality}}$ contains quality flags, $\mathcal{S}_{\mathrm{summary}}$ contains optional descriptive outputs, and $\mathcal{U}_{\mathrm{unresolved}}$ identifies records requiring manual review.

\subsection{Data Modeling Agent}
\label{sec:data_modeling_agent}

The Data Modeling Agent constructs, estimates, and evaluates behavioral models based on the processed data and the user instruction $\mathbf{u}_{\mathrm{mod}}$. The instruction may specify the prediction target, explanatory variables, candidate model families, validation scheme, evaluation metrics, and required outputs. The agent may use an LLM to translate the instruction into a formal model specification or executable code, but numerical estimation is delegated to the selected statistical or machine-learning tool.

A modeling request is represented as:

\begin{equation}
\mathcal{G}_{\mathrm{mod}}
=
\left(
\mathcal{Y},
\mathcal{X},
\mathbb{M},
\mathcal{V},
\mathcal{C}
\right)
\label{eq:modeling_request}
\end{equation}

\noindent where $\mathcal{Y}$ defines the target outcome, $\mathcal{X}$ defines the candidate predictors, $\mathbb{M}$ is the authorized model library, $\mathcal{V}$ defines the validation design and performance metrics, and $\mathcal{C}$ contains modeling constraints. The candidate library may include random-utility models, statistical classifiers, machine-learning models, neural networks, and LLM-based role-play predictors.

Let $\mathcal{D}_{\mathrm{tr}}$, $\mathcal{D}_{\mathrm{va}}$, and $\mathcal{D}_{\mathrm{te}}$ denote the training, validation, and test partitions. The agent selects a model family $m$ and configuration $\boldsymbol{\lambda}$ by solving:

\begin{equation}
\left(
m^{\star},
\boldsymbol{\lambda}^{\star}
\right)
\in
\arg\min_{
m\in\mathbb{M},
\boldsymbol{\lambda}\in\Lambda_m
}
\left\{
\widehat{\mathcal{R}}_{\mathrm{va}}
\left(
m,
\boldsymbol{\lambda};
\mathcal{D}_{\mathrm{tr}},
\mathcal{D}_{\mathrm{va}}
\right)
+
\eta
\Omega
\left(
m,
\boldsymbol{\lambda}
\right)
\right\}
\label{eq:model_selection}
\end{equation}

\noindent where $\widehat{\mathcal{R}}_{\mathrm{va}}$ is the validation loss, $\Omega(\cdot)$ is an optional complexity or constraint penalty, and $\eta$ controls the penalty weight. The selected model is then fitted as:

\begin{equation}
\widehat{\mathcal{M}}
=
\operatorname{Train}
\left(
m^{\star},
\boldsymbol{\lambda}^{\star};
\mathcal{D}_{\mathrm{tr}}
\cup
\mathcal{D}_{\mathrm{va}}
\right)
\label{eq:final_model_training}
\end{equation}

For commuter mode choice, the general prediction function can be written as:

\begin{equation}
\widehat{y}_{is}
=
f_{\widehat{\boldsymbol{\theta}}}
\left(
\mathbf{p}_{i},
\mathbf{z}_{is},
\mathbf{a}_{is}
\right)
\label{eq:general_mode_prediction}
\end{equation}

\noindent where $\mathbf{p}_{i}$ contains individual characteristics, $\mathbf{z}_{is}$ contains scenario-specific attributes, $\mathbf{a}_{is}$ defines the available choice set, and $\widehat{\boldsymbol{\theta}}$ denotes the fitted parameters or learned model state.

When a random-utility model is selected, the utility of mode $j$ may be represented as:

\begin{equation}
U_{isj}
=
V_{isj}
+
\varepsilon_{isj},
\qquad
V_{isj}
=
\boldsymbol{\beta}^{\top}
\mathbf{x}_{isj}
\label{eq:random_utility}
\end{equation}

\noindent where $\mathbf{x}_{isj}$ contains individual-, scenario-, and alternative-specific variables. Under a multinomial logit specification, the choice probability is:

\begin{equation}
P_{isj}
=
\frac{
a_{isj}\exp\left(V_{isj}\right)
}{
\sum_{k\in\mathcal{J}}
a_{isk}\exp\left(V_{isk}\right)
}
\label{eq:mnl_probability}
\end{equation}

\noindent where $\mathcal{J}$ is the set of travel modes and $a_{isj}\in\{0,1\}$ indicates whether mode $j$ is available \citep{train2009}. More flexible specifications, such as mixed logit or latent-class models, may be selected when the user requests explicit treatment of taste heterogeneity \citep{mcfadden2000mixed}.

The framework also permits LLM-based choice prediction through alternative prompt framings. Let $\rho(\cdot)$ denote a prompt-construction function and $\psi$ denote the LLM configuration. For repeated run $r$, the predicted choice is:

\begin{equation}
\widehat{y}_{is}^{(r)}
=
F_{\psi}
\left[
\rho
\left(
\mathbf{p}_{i},
\mathbf{z}_{is},
\mathbf{a}_{is},
\mathbf{u}_{\mathrm{mod}}
\right);
\epsilon_{r}
\right],
\qquad
r=1,\ldots,R
\label{eq:llm_roleplay}
\end{equation}

\noindent where $\epsilon_{r}$ represents stochastic variation across repeated runs. The empirical LLM-based probability of choosing mode $j$ is:

\begin{equation}
\widehat{P}_{isj}^{\mathrm{LLM}}
=
\frac{1}{R}
\sum_{r=1}^{R}
\mathbb{I}
\left(
\widehat{y}_{is}^{(r)}=j
\right)
\label{eq:llm_choice_probability}
\end{equation}

The LLM predictor may operate in zero-shot or few-shot mode. In zero-shot prediction, the prompt contains the respondent profile, commuting scenario, available modes, and output requirements but no labeled examples. In few-shot prediction, labeled examples are added from the training data. Examples from validation or test respondents are not permitted.

When each respondent contributes multiple scenarios, data partitioning is conducted at the respondent level rather than the record level. Therefore,

\begin{equation}
\mathcal{I}_{\mathrm{tr}}
\cap
\mathcal{I}_{\mathrm{va}}
=
\mathcal{I}_{\mathrm{tr}}
\cap
\mathcal{I}_{\mathrm{te}}
=
\mathcal{I}_{\mathrm{va}}
\cap
\mathcal{I}_{\mathrm{te}}
=
\varnothing
\label{eq:respondent_split}
\end{equation}

\noindent where $\mathcal{I}_{\mathrm{tr}}$, $\mathcal{I}_{\mathrm{va}}$, and $\mathcal{I}_{\mathrm{te}}$ are the respondent sets assigned to the three partitions. This rule prevents information from the same respondent from appearing in both model development and evaluation.

The modeling output is represented as:

\begin{equation}
\mathcal{R}_{\mathrm{mod}}
=
\left(
\widehat{\boldsymbol{\theta}},
\widehat{\mathbf{Y}},
\mathcal{I}_{\mathrm{effect}},
\mathcal{U}_{\mathrm{model}},
\mathcal{G}_{\mathrm{diagnostic}},
\mathcal{V}_{\mathrm{evaluation}}
\right)
\label{eq:modeling_outputs}
\end{equation}

\noindent where $\widehat{\boldsymbol{\theta}}$ contains the estimated parameters or fitted model state, $\widehat{\mathbf{Y}}$ contains predictions, $\mathcal{I}_{\mathrm{effect}}$ contains effect estimates or variable importance measures, $\mathcal{U}_{\mathrm{model}}$ contains uncertainty information, $\mathcal{G}_{\mathrm{diagnostic}}$ contains diagnostic outputs, and $\mathcal{V}_{\mathrm{evaluation}}$ contains validation metrics. The agent must distinguish predictive associations from causal effects unless the study design supports causal identification.

Each workflow version follows a forward execution path, with each stage producing a structured output that the next agent consumes through a defined interface: a survey specification is provided to the Data Collection Agent, which returns raw data; the raw data are provided to the Data Processing Agent, which returns a processed dataset; and the processed dataset is provided to the Data Modeling Agent, which returns fitted models, predictions, diagnostics, and evaluation metrics. As shown by the dashed arrows in Figure~\ref{fig:framework}, diagnostics from one stage may motivate a researcher-approved revision of an earlier survey, processing, or modeling specification in a subsequent workflow version. The present case study implemented one formal forward pass; the dashed paths represent supported refinement mechanisms for subsequent versions. Section~4 describes how the three agents were instantiated for the McGill commuter mode-choice case study, and Section~5 reports the resulting analyses and predictions.


\section{Case Study}

The framework in Section 3 was instantiated on a commuter mode choice case study at McGill University in Montreal, Canada, a setting where students commute by walking, cycling, transit, and driving across a full range of seasonal weather, from summer heat to winter snow. This section describes how each of the three agents was implemented for that case study; detailed results appear in Section 5.

\subsection{Data Collection Agent: Chatbot Survey and Deployment}

\begin{figure}[t]
\centering
\begin{subfigure}[t]{0.4\linewidth}
    \centering
    \includegraphics[width=\linewidth]{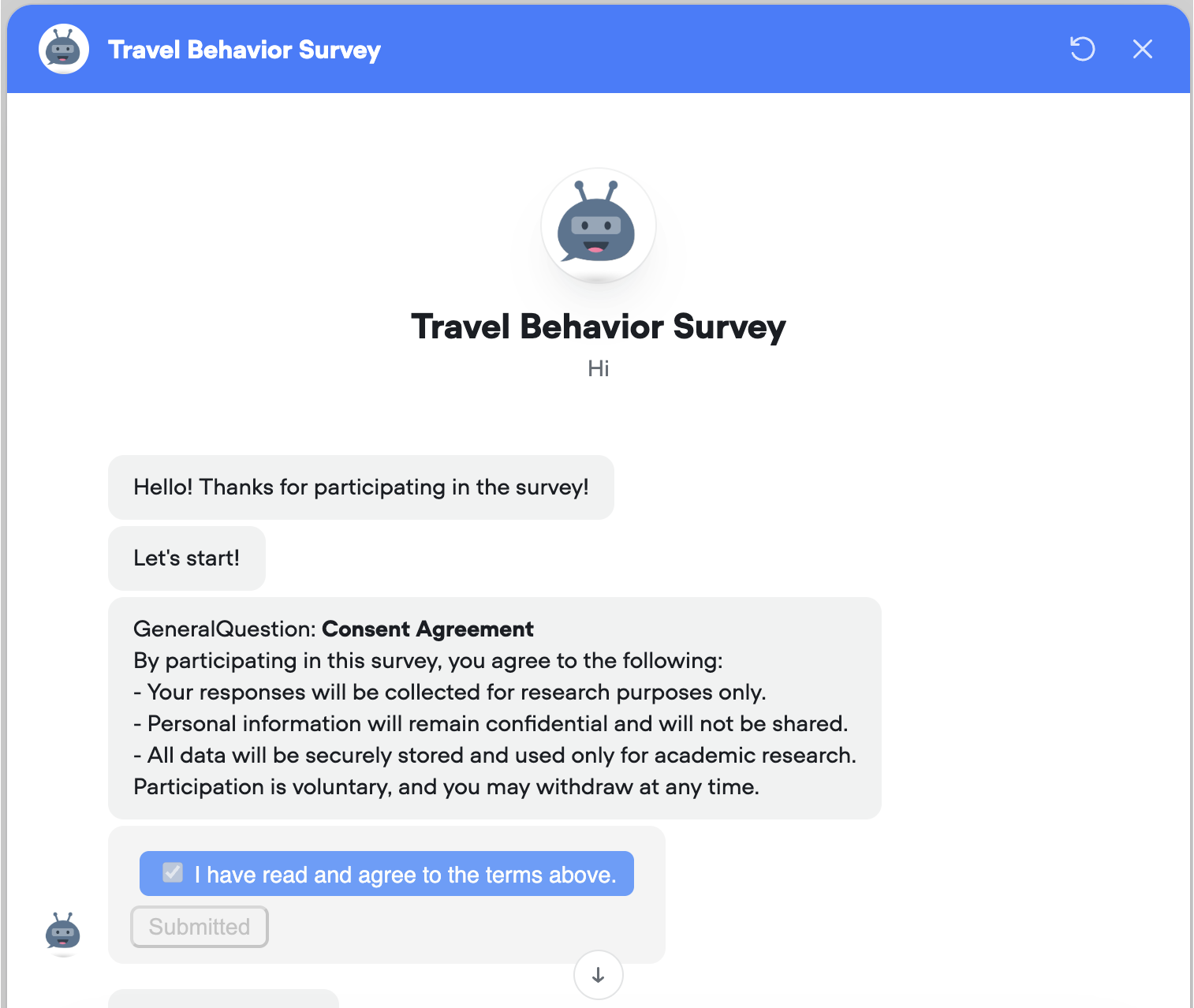}
    \label{fig:chatbot1}
\end{subfigure}\hspace{0.01\linewidth}
\begin{subfigure}[t]{0.4\linewidth}
    \centering
    \includegraphics[width=\linewidth]{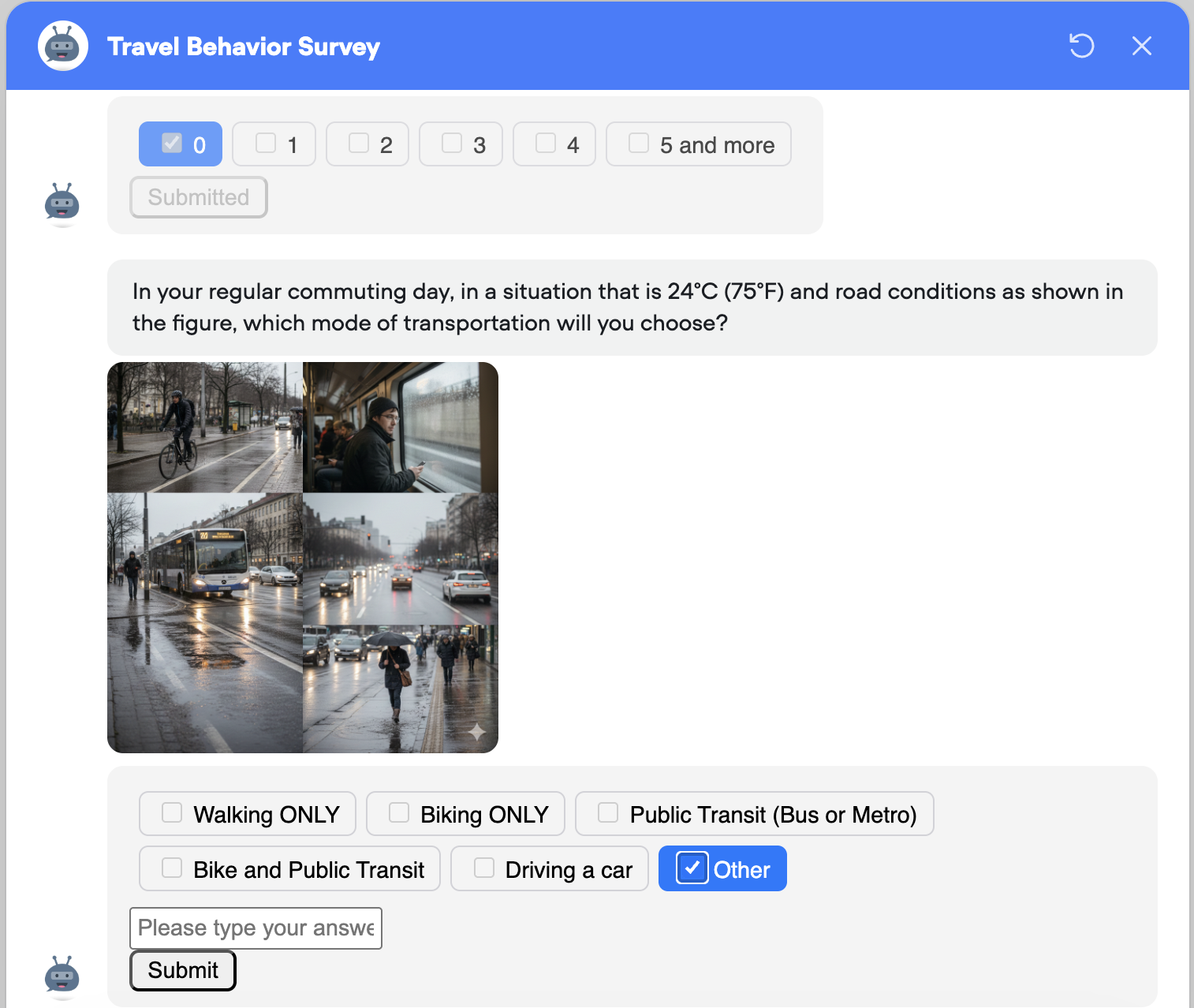}
    \label{fig:chatbot2}
\end{subfigure}

\caption{Chatbot survey interface and sample weather-scenario question with embedded generated images.}
\label{fig:chatbot}
\end{figure}

The Data Collection Agent was implemented as a purpose-built conversational chatbot, \emph{TravelBehaviorSurveyBot}, developed in Voiceflow and deployed as a web application accessible through a public link and QR code (Figure~\ref{fig:chatbot}). Respondents completed the survey on their own devices through a guided dialogue that collected demographic and mobility-resource information, usual summer and winter travel patterns, weekly mode-use frequencies, and ratings of factors influencing mode choice, followed by five SP weather scenarios. Each scenario presented one photorealistic image generated using defined prompts to represent Sunny, Hot--humid, Rainy, Foggy/cold, and Snowy commuting conditions. All respondents were shown the same fixed image for each scenario. For each scenario, respondents selected among five modeled alternatives: walking, cycling, public transit, bike and public transit combined, and driving. The chatbot followed predefined survey logic and did not alter the experimental attributes or choice set during deployment. Figure~\ref{fig:chatbot} illustrates the chatbot interface and the weather images presented during the survey. This design required neither app installation nor an interviewer and ensured consistent presentation of the scenarios across respondents. The Phase~1 sample included 92 McGill student commuters.

\subsection{Data Processing Agent}

The Data Processing Agent converted the chatbot's raw, JSON-encoded response export into an analysis-ready dataset. Processing steps included parsing nested list- and slider-valued fields, validating and recoding categorical responses, constructing derived features, and flagging incomplete or inconsistent records for exclusion. Of the 460 potential observations from 92 respondents and five scenarios, six were excluded because the recorded mode choice was missing or unparseable. The final processed dataset therefore contained 454 valid respondent--scenario observations. Key derived features include weekly mode-use frequency by season, the seven factor-importance scores, binary vehicle-, license-, and bicycle-ownership flags, and years of cycling experience, which are reported in Section~5.1.

\subsection{Data Modeling Agent: Modeling and LLM Setup}

The Data Modeling Agent implemented two families of models on the processed dataset. The traditional family comprises an MNL estimated in Biogeme \citep{bierlaire2003biogeme} and ML classifiers, principally logistic regression and random forest. The LLM family comprises nine locally run models ranging from 2 to 35 billion parameters. The traditional and LLM approaches were executed as parallel model families rather than as sequential modeling stages. For each respondent--scenario pair, an LLM receives a natural-language representation of the respondent profile together with the corresponding weather condition and is asked to predict the selected travel mode. The prompting experiments follow a structured progression in which two prompt framings, Expert (EXP) and Role-Play (RP), are crossed with two levels of contextual information, Base Context (BC) and Richer Context (RC); the richer setting incorporates additional habitual travel information from respondents' profiles. These four base conditions are then extended through persona-augmented prompting based on respondent personas extracted using a three-class latent class analysis, few-shot in-context learning with labeled examples at several values of $k$, and direct image input for vision-capable models using the same weather images shown to survey respondents. Predictions are evaluated for both the five-alternative mode-choice task and a binary active-versus-non-active classification, with all LLM and traditional models assessed on the same 454 respondent--scenario observations. Accuracy is reported and interpreted relative to random-guessing baselines of 20.0\% for the five-class task and 50.0\% for the binary task. The full traditional-model estimates and LLM experiments are reported in Section~5.

\section{Results}
\label{sec:results}

\subsection{Sample Profile and Traditional Model Results}

\subsubsection{Sample Profile and Weather-Related Mode Choice}

\begin{figure}[t]
\centering
\includegraphics[width=0.95\linewidth]{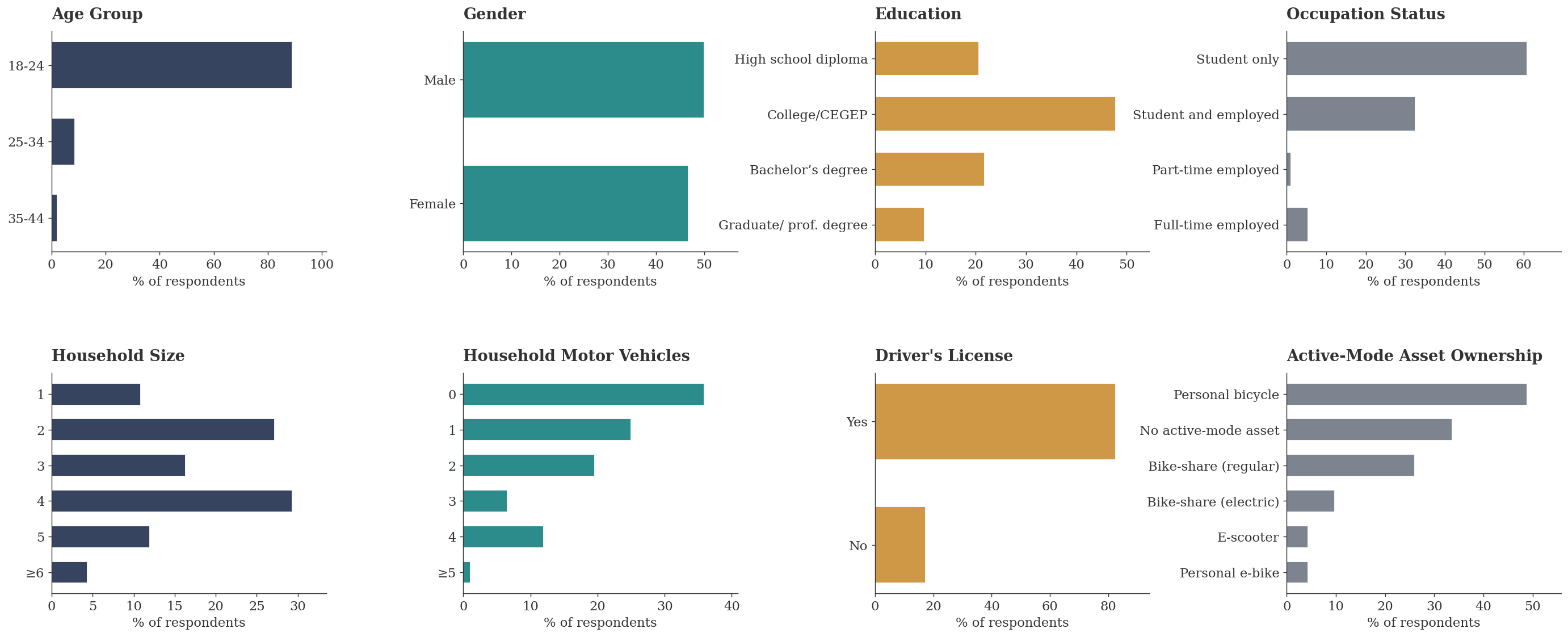}
\caption{Sample profile of survey respondents.}
\label{fig:sampleprofile}
\end{figure}

The Phase~1 sample comprises 92 McGill student commuters, predominantly aged 18--24, with a nearly balanced gender distribution and heterogeneous mobility resources (Figure~\ref{fig:sampleprofile}). Although 83\% of respondents hold a driver's license, 36\% live in households without a motor vehicle and approximately half own a personal bicycle. Self-reported habitual commute patterns also vary considerably by season: cycling declines from 13.0\% of primary summer commutes to 2.2\% in winter, while public transit increases from 20.7\% to 43.5\%, indicating substantial seasonal adaptation in travel behavior. A three-class latent class analysis further identifies a small group of year-round cyclists and two larger groups characterized by active--transit and car--transit travel patterns; these behavioral personas are subsequently used in the persona-based LLM experiments.

\begin{figure}[t]
\centering
\includegraphics[width=0.95\linewidth]{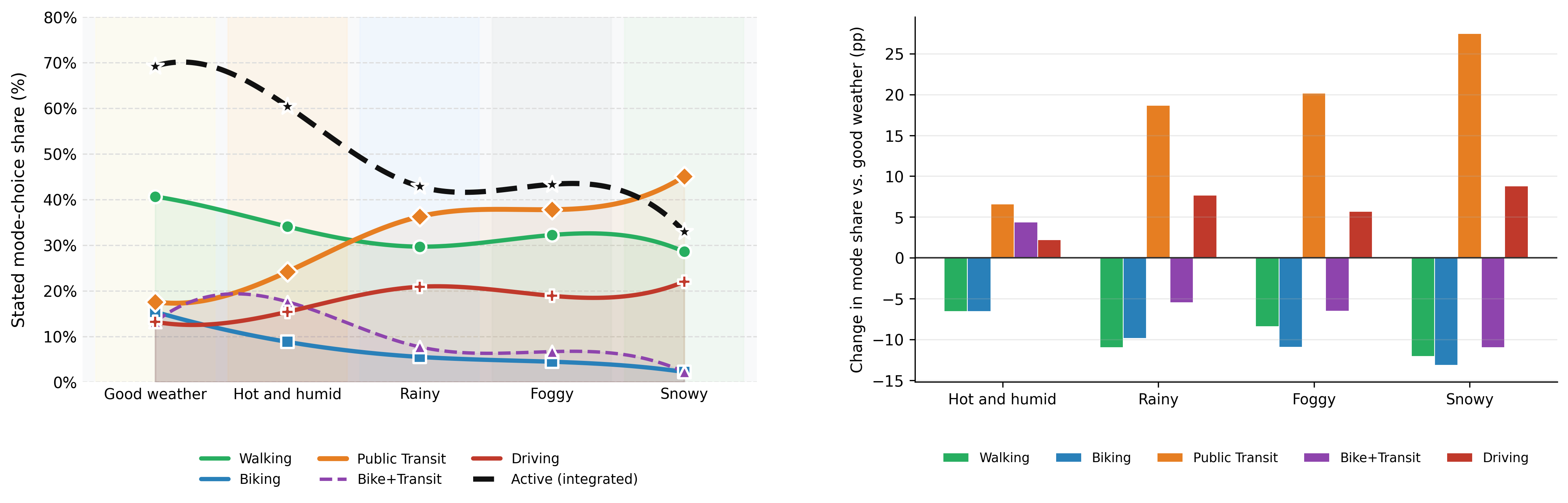}
\caption{Stated mode share across weather scenarios (left) and change in mode share
relative to the Sunny scenario, in percentage points (right).}
\label{fig:modeshare}
\end{figure}

The stated-preference choices vary systematically across the five weather scenarios (Figure~\ref{fig:modeshare}). Walking declines moderately across conditions, whereas cycling and bike+transit show the largest reductions and approach minimal shares under the Snowy scenario. In contrast, public transit gains share, increasing from 17.6\% under the Sunny scenario to 45.1\% under the Snowy scenario, while driving changes comparatively little. Overall, the descriptive results indicate that active and multimodal alternatives are more sensitive to the predefined scenarios than the motorized alternatives, with public transit serving as the primary substitute as conditions become less favorable.

\subsubsection{Discrete Choice and Machine-Learning Models}

The MNL model provides an interpretable benchmark for the weather-related mode-choice patterns observed descriptively. The expanded specification includes weather-scenario indicators, mobility resources, cycling experience, age, household income, and self-rated weather importance, and significantly improves model fit relative to the baseline specification (LR $=53.92$, $df=8$, $p<0.0001$). With Walking and Sunny weather as the reference categories, the coefficients indicate lower relative utility for cycling and higher relative utility for public transit and driving under adverse scenarios. The Cycling coefficients are negative under Rainy ($\beta=-0.856$, $p=0.044$), Foggy/cold ($\beta=-1.166$, $p=0.012$), and Snowy conditions ($\beta=-1.820$, $p=0.009$), with the largest reduction under the Snowy scenario. Bike+Transit also has a negative coefficient under Snowy conditions ($\beta=-1.695$, $p=0.050$). In contrast, Public Transit has positive coefficients under Hot--humid ($\beta=0.518$, $p=0.030$), Rainy ($\beta=1.070$, $p<0.001$), Foggy/cold ($\beta=1.024$, $p<0.001$), and Snowy conditions ($\beta=1.324$, $p<0.001$). Driving similarly has positive coefficients under Rainy ($\beta=0.949$, $p=0.007$), Foggy/cold ($\beta=0.723$, $p=0.031$), and Snowy conditions ($\beta=1.042$, $p=0.003$). Bicycle ownership and cycling experience are positively associated with the cycling-related alternatives, whereas household motor-vehicle availability and holding a driver's license are positively associated with driving. Overall, the MNL estimates are consistent with the descriptive results in Figure~\ref{fig:modeshare}.

\begin{table}[t]
\centering
\caption{Selected Multinomial Logit Coefficients
(Reference Mode: Walking; Reference Weather: Sunny)}
\label{tab:mnl}
\footnotesize
\begin{tabular}{@{}llrrr@{}}
\toprule
\textbf{Mode} & \textbf{Covariate} & \textbf{Coef.} &
\textbf{Rob. \emph{t}} & \textbf{\emph{p}} \\
\midrule
Cycling & Rainy (vs. Sunny) & $-$0.856 & $-$2.01 & 0.044* \\
Cycling & Foggy/cold (vs. Sunny) & $-$1.166 & $-$2.52 & 0.012* \\
Cycling & Snowy (vs. Sunny) & $-$1.820 & $-$2.63 & 0.009** \\
Cycling & Owns bicycle & 1.219 & 2.06 & 0.039* \\
Cycling & Cycling experience (yrs.) & 0.341 & 2.67 & 0.008** \\
Public Transit & Hot--humid (vs. Sunny) & 0.518 & 2.17 & 0.030* \\
Public Transit & Rainy (vs. Sunny) & 1.070 & 3.68 & \textless 0.001*** \\
Public Transit & Foggy/cold (vs. Sunny) & 1.024 & 4.08 & \textless 0.001*** \\
Public Transit & Snowy (vs. Sunny) & 1.324 & 4.59 & \textless 0.001*** \\
Bike+Transit & Snowy (vs. Sunny) & $-$1.695 & $-$1.96 & 0.050* \\
Bike+Transit & Owns bicycle & 1.572 & 2.64 & 0.008** \\
Bike+Transit & Cycling experience (yrs.) & 0.281 & 2.49 & 0.013* \\
Bike+Transit & Self-rated weather importance & 0.471 & 2.54 & 0.011* \\
Driving & Rainy (vs. Sunny) & 0.949 & 2.68 & 0.007** \\
Driving & Foggy/cold (vs. Sunny) & 0.723 & 2.16 & 0.031* \\
Driving & Snowy (vs. Sunny) & 1.042 & 3.01 & 0.003** \\
Driving & Household motor vehicles & 0.767 & 2.67 & 0.007** \\
Driving & Driver's license & 1.769 & 2.13 & 0.033* \\
\bottomrule
\multicolumn{5}{l}{\footnotesize
* $p\leq0.05$; ** $p<0.01$; *** $p<0.001$.}
\end{tabular}
\end{table}

For predictive benchmarking, the MNL, logistic regression, and random forest were evaluated using the same five-fold cross-validation procedure grouped by respondent, ensuring that observations from the same individual did not appear in both training and test sets. Random forest produced the highest accuracy, reaching 69.6\% for the five-class mode-choice task and 88.8\% for the binary active-versus-non-active task. Logistic regression reached 60.2\% and 85.1\%, respectively, while the MNL achieved 44.7\% five-class and 81.2\% binary accuracy. The five-class results show a predictive advantage for the machine-learning models, particularly random forest, whereas the smaller differences in binary accuracy should be interpreted in light of the strong majority-class imbalance. These findings highlight complementary roles: the MNL provides behaviorally interpretable evidence on weather scenarios and traveler characteristics, whereas the ML models provide stronger data-trained benchmarks for the LLM experiments that follow.

\subsection{LLM Zero-Shot}

The zero-shot experiments evaluate two dimensions of prompt design, context richness and prompt framing. Base Context (BC) includes respondent demographics, mobility resources, factor-importance ratings, and the weather scenario, whereas Richer Context (RC) additionally incorporates additional habitual travel information. Expert (EXP) framing asks the LLM to predict the respondent's choice as a transportation expert, while Role-Play (RP) framing asks the model to simulate the respondent directly. The four resulting conditions, EXP-RC, EXP-BC, RP-RC, and RP-BC, were evaluated across nine locally deployed LLMs ranging from 2 to 35 billion parameters (Table~\ref{tab:llmall}).

\begin{figure}[t]
\centering
\includegraphics[width=0.95\linewidth]{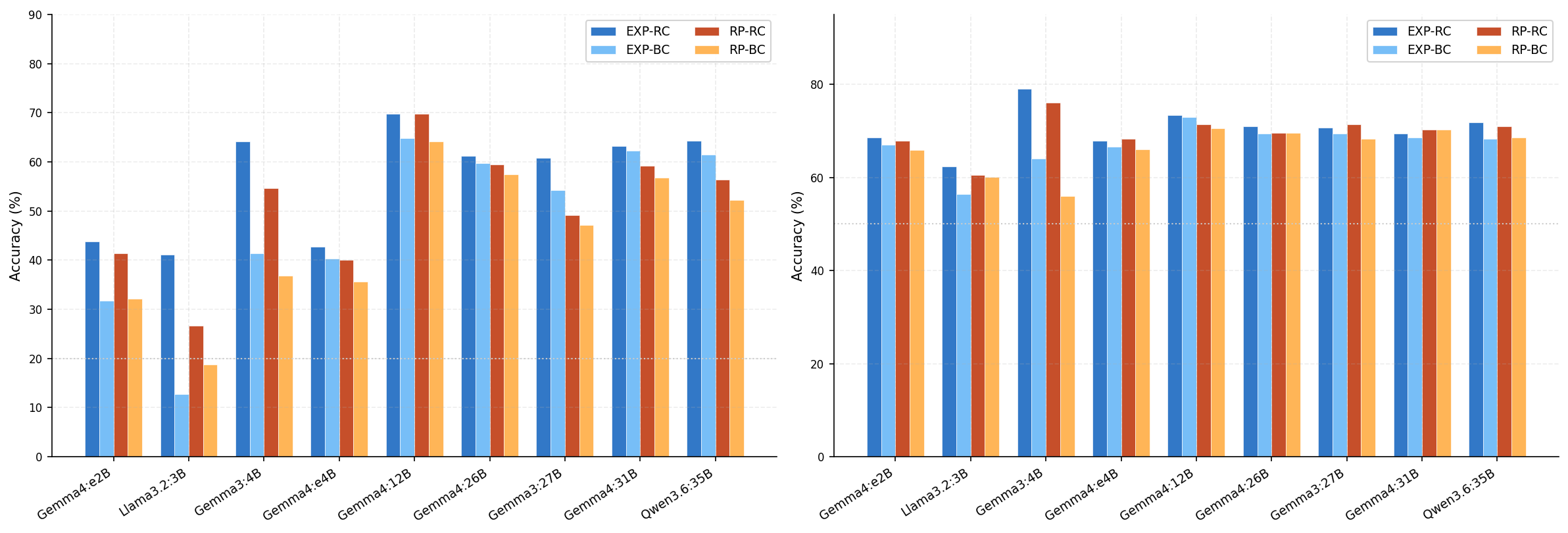}
\caption{Zero-shot accuracy across all four base conditions (Expert/Role-Play $\times$ Base/Richer Context) for nine models, shown separately for five-class and binary accuracy.}
\label{fig:zeroshot}
\end{figure}

Figure~\ref{fig:zeroshot} reveals two consistent patterns. First, incorporating additional habitual travel information substantially improves zero-shot prediction. For Gemma~3:4B, five-class accuracy increases from 41.3\% under EXP-BC to 64.2\% under EXP-RC, indicating that habitual travel information provides a strong predictive signal beyond demographics and mobility resources. Second, Expert framing generally outperforms Role-Play framing at the same context level, with the difference more pronounced among smaller models. Larger models in the evaluated set generally produce higher five-class accuracy than the smallest models, although the pattern is not monotonic and model size is confounded with model family and architecture. The highest zero-shot five-class point estimate is achieved by Gemma~4:12B at approximately 70\%, whereas binary accuracy varies less across models. Together, these results indicate that the information supplied to the model and the framing of the prediction task can be as important as the selected model.

\begin{table}[t]
\centering
\caption{LLM Zero-Shot Five-Class and Binary Accuracy}
\label{tab:llmall}
\scriptsize
\begin{tabular}{@{}lcccccccc@{}}
\toprule
& \multicolumn{2}{c}{\textbf{EXP-RC}} & \multicolumn{2}{c}{\textbf{EXP-BC}} & \multicolumn{2}{c}{\textbf{RP-RC}} & \multicolumn{2}{c}{\textbf{RP-BC}} \\
\cmidrule(lr){2-3}\cmidrule(lr){4-5}\cmidrule(lr){6-7}\cmidrule(lr){8-9}
\textbf{Model} & 5-cl. & Binary & 5-cl. & Binary & 5-cl. & Binary & 5-cl. & Binary \\
\midrule
Gemma 4:e2B & 44.0 $\pm$ 4.5 & 68.7 $\pm$ 4.1 & 31.6 $\pm$ 4.1 & 66.8 $\pm$ 4.3 & 41.2 $\pm$ 5.0 & 67.9 $\pm$ 4.2 & 32.3 $\pm$ 4.3 & 65.8 $\pm$ 4.1 \\
Llama 3.2:3B & 41.2 $\pm$ 4.3 & 62.5 $\pm$ 4.5 & 12.7 $\pm$ 3.6 & 56.7 $\pm$ 4.4 & 26.7 $\pm$ 3.7 & 60.7 $\pm$ 4.3 & 18.6 $\pm$ 3.6 & 60.3 $\pm$ 4.5 \\
Gemma 3:4B & 64.2 $\pm$ 4.4 & 79.1 $\pm$ 3.3 & 41.3 $\pm$ 4.0 & 64.1 $\pm$ 4.4 & 54.5 $\pm$ 4.2 & 75.9 $\pm$ 3.9 & 36.7 $\pm$ 4.2 & 55.6 $\pm$ 4.4 \\
Gemma 4:e4B & 42.8 $\pm$ 4.4 & 67.9 $\pm$ 4.4 & 40.4 $\pm$ 4.5 & 66.4 $\pm$ 4.2 & 40.1 $\pm$ 4.5 & 68.3 $\pm$ 4.0 & 35.7 $\pm$ 4.5 & 66.0 $\pm$ 4.2 \\
Gemma 4:12B & 69.9 $\pm$ 3.9 & 73.5 $\pm$ 4.2 & 64.6 $\pm$ 4.3 & 73.0 $\pm$ 4.2 & 69.8 $\pm$ 4.4 & 71.5 $\pm$ 4.1 & 63.9 $\pm$ 5.0 & 70.6 $\pm$ 4.1 \\
Gemma 4:26B & 61.2 $\pm$ 4.6 & 71.1 $\pm$ 4.0 & 59.6 $\pm$ 4.7 & 69.6 $\pm$ 4.4 & 59.3 $\pm$ 4.5 & 69.7 $\pm$ 4.0 & 57.4 $\pm$ 4.3 & 69.7 $\pm$ 4.5 \\
Gemma 3:27B & 60.7 $\pm$ 4.6 & 71.0 $\pm$ 4.1 & 54.1 $\pm$ 4.1 & 69.5 $\pm$ 4.2 & 49.1 $\pm$ 4.5 & 71.5 $\pm$ 4.2 & 47.1 $\pm$ 4.6 & 68.3 $\pm$ 4.2 \\
Gemma 4:31B & 63.1 $\pm$ 4.4 & 69.5 $\pm$ 4.0 & 62.2 $\pm$ 4.7 & 68.6 $\pm$ 4.0 & 59.2 $\pm$ 4.6 & 70.4 $\pm$ 4.0 & 56.7 $\pm$ 4.9 & 70.3 $\pm$ 4.3 \\
Qwen 3.6:35B & 64.3 $\pm$ 4.3 & 71.9 $\pm$ 4.0 & 61.5 $\pm$ 4.4 & 68.4 $\pm$ 4.1 & 56.4 $\pm$ 4.6 & 71.0 $\pm$ 4.3 & 52.1 $\pm$ 4.7 & 68.5 $\pm$ 4.2 \\
\bottomrule
\end{tabular}
\end{table}

\begin{figure}[t]
    \centering
    \includegraphics[width=0.98\linewidth]
        {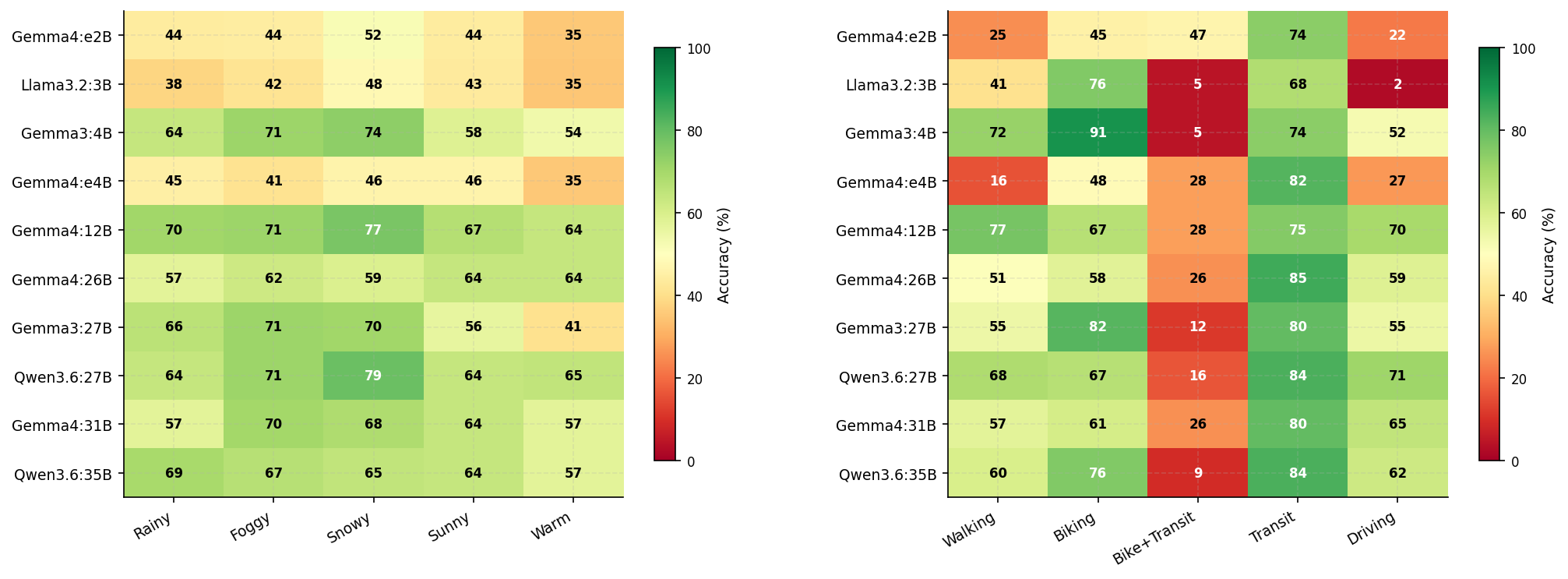}
    \par\vspace{0.3em}
    \includegraphics[width=0.98\linewidth]
        {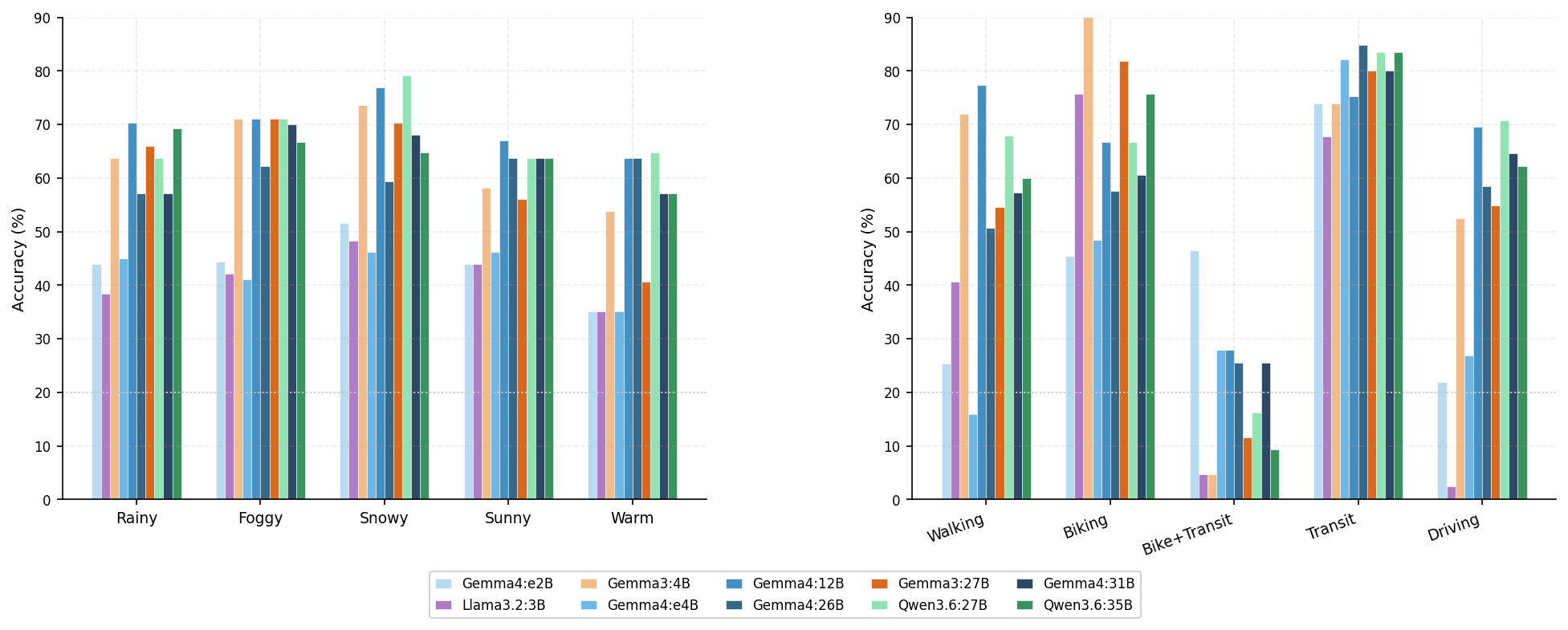}
    \caption{Accuracy patterns by weather condition and mode across all evaluated models, with a bar-chart breakdown of weather- and mode-level accuracy under the EXP-RC experiment.}
    \label{fig:heatmap}
\end{figure}

Prediction difficulty also varies systematically across weather conditions and travel modes. As shown in Figure~\ref{fig:heatmap}, Snowy and Rainy scenarios are generally easier to predict because they produce stronger and more consistent shifts toward public transit and away from cycling. In contrast, Sunny and Warm conditions yield more heterogeneous choices and lower prediction accuracy. At the mode level, Public Transit is predicted relatively well across models, whereas Bike+Transit is consistently the most difficult alternative. These differences suggest that LLM accuracy depends not only on model and prompt configuration but also on the behavioral regularity of the scenario and alternative being predicted.

\subsection{Advanced LLM Configurations: Persona, Few-Shot, and Vision}

The advanced experiments examine whether additional behavioral and contextual information improves prediction beyond the base zero-shot configurations. Persona augmentation is most beneficial when explicit travel history is unavailable. Under Base Context, adding a behavioral persona substantially improves some smaller models, particularly Llama~3.2:3B, whereas its effect is limited when Richer Context already contains respondents' habitual seasonal travel modes. This pattern suggests that persona information mainly summarizes behavioral information when direct travel-history information is absent, rather than adding substantial information once habitual travel information is already available.

\begin{figure}[t]
\centering
\includegraphics[width=\linewidth]{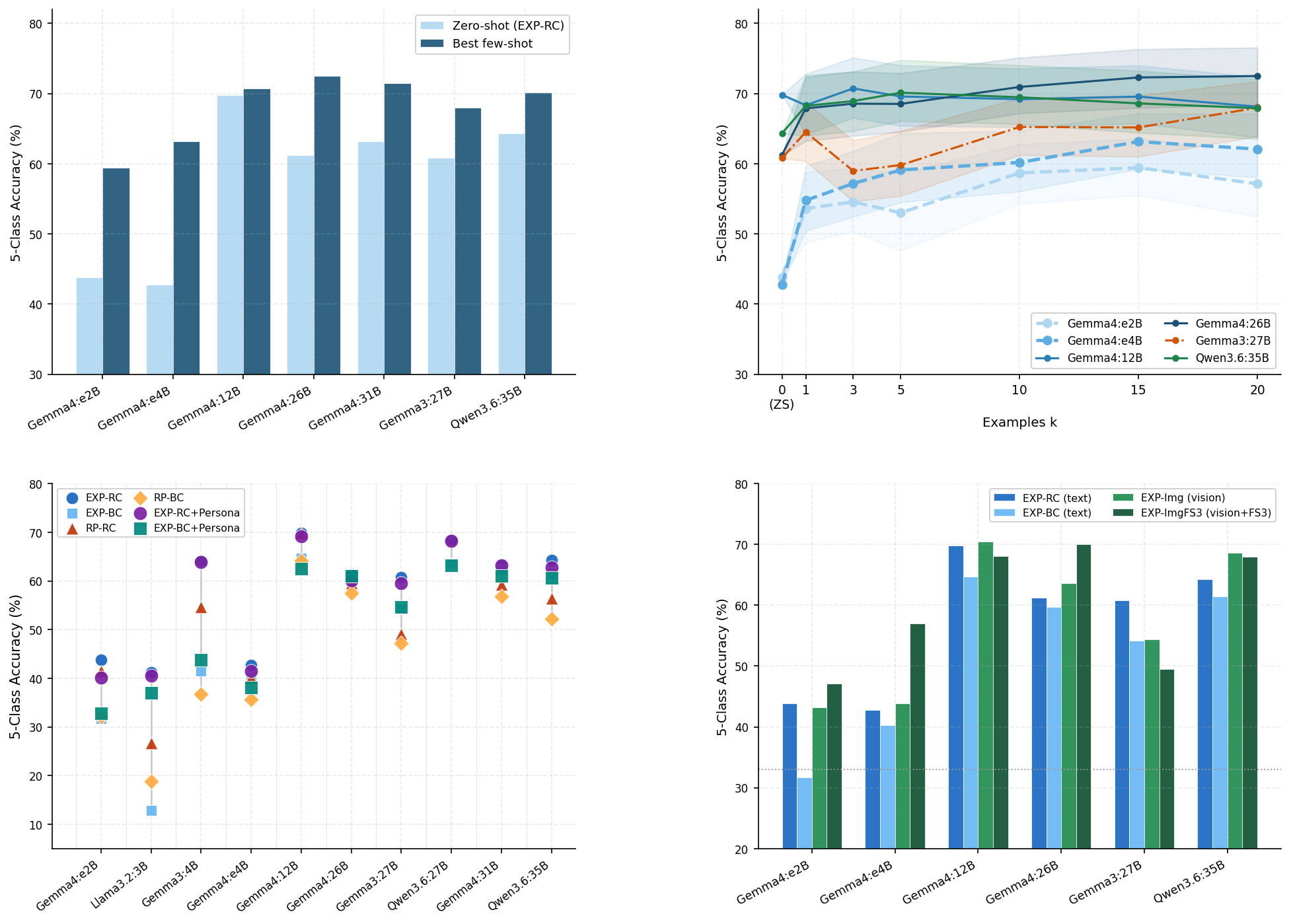}
\caption{Combined LLM results. Top row: (left) zero-shot versus peak few-shot
five-class accuracy per model, and (right)
few-shot scaling curves across the six models. Bottom row: (left) experiment spread across all conditions, and (right) vision augmentation results comparing
text-only baselines (EXP-RC and EXP-BC) against vision input (EXP-Img) and
vision with three-shot prompting (EXP-ImgFS3).}
\label{fig:combined}
\end{figure}

Few-shot prompting provides further improvements in five-class prediction, although the magnitude of the gain varies across models and diminishes as additional examples are introduced. Figure~\ref{fig:combined} shows that a small number of labeled examples improves performance for most models, with proportionally larger gains among smaller models. Accuracy generally stabilizes after approximately ten examples, while additional demonstrations provide limited or inconsistent improvements. The alternative example-selection strategies considered produce broadly comparable performance, suggesting that the presence of informative demonstrations is more important than a complex retrieval strategy. These improvements are more evident for the five-class task than for the binary active-versus-non-active classification, indicating that few-shot prompting is particularly effective for predicting specific travel modes.

Vision-capable models were further evaluated using the same generated weather images presented to survey respondents. Visual inputs provide additional contextual information for several models, although the effect is not uniform. For some larger models, vision-based prompting approaches or exceeds the corresponding text-based Richer Context performance, indicating that the specific scenario images provide predictive signal that can complement or partially substitute for text-based contextual descriptions. Qwen~3.6:35B, for example, improves when the textual weather representation is replaced by the scenario image, while Gemma~4:26B benefits further when vision input is combined with few-shot examples. Other models show smaller gains or declines, indicating that multimodal capability alone does not guarantee improved mode-choice prediction.

Overall, the advanced experiments indicate that additional information is most valuable when it contributes behavioral or contextual signals not already represented in the prompt. Habitual travel information remains the most consistent source of predictive information, persona descriptions are particularly useful when that history is unavailable, and few-shot examples provide useful task guidance with rapidly diminishing returns. Vision inputs provide a distinct source of contextual information and can improve prediction for selected models, supporting further evaluation of multimodal LLMs in travel-behavior applications.

\section{Discussion}

The results demonstrate how conversational data collection, conventional behavioral models, and LLM-based prediction can be connected within a common multi-agent workflow for travel-behavior research. Beyond the performance of any individual model, the study provides evidence on three broader questions: whether an AI-assisted survey can capture meaningful context-dependent travel behavior, which forms of information are most useful for LLM-based mode-choice prediction, and how these models compare with established DCMs and ML approaches.

\textbf{Multi-agent workflow and behavioral evidence.}
The workflow links the Data Collection, Data Processing, and Data Modeling Agents through structured outputs, creating a transparent and modular process from survey administration to behavioral prediction. This organization does not improve accuracy by itself; rather, it enables consistent data transfer and systematic evaluation of where performance gains arise. The chatbot survey produced coherent differences across the weather scenarios: cycling and bike+transit declined as conditions worsened, whereas public transit gained substantial share. The MNL estimates were consistent with these descriptive patterns, showing lower relative utility for cycling-related modes and higher relative utility for transit and driving under adverse scenarios. This agreement supports the internal coherence of the survey responses, although it does not by itself establish external validity or actual travel behavior under observed weather.

\textbf{LLM configuration and comparison with conventional models.}
LLM performance was strongly influenced by the information included in the prompt. RC, which adds habitual travel information to BC, produced the most consistent improvement, showing that habitual travel behavior contains predictive information not fully captured by demographics and mobility resources. EXP framing generally outperformed RP, particularly for smaller models, while gains from increasing model size diminished among larger models.

Persona augmentation and few-shot prompting were most useful when they supplied missing information. Persona descriptions improved prediction mainly under BC, where habitual travel information was unavailable, but added little under RC. Few-shot examples improved five-class accuracy for several models, particularly smaller ones, although the gains generally stabilized after a small number of demonstrations.

The conventional and LLM-based models therefore play complementary roles. The MNL provides behavioral interpretation, while logistic regression and random forest offer data-trained predictive benchmarks. Random forest achieved the highest trained-model accuracy, while the best zero-shot LLM produced a comparable five-class point estimate without task-specific fitting to the survey records. This result suggests potential value for LLM-based prediction when labeled travel-behavior data are limited, although the small sample and multiple evaluated configurations require cautious interpretation. The five-class task provides a clearer assessment of whether models distinguish among individual modes.

\textbf{Vision-based prediction.}
The vision experiments directly connect the survey and prediction stages of the multi-agent workflow by providing the models with the same generated weather images shown to respondents. The best vision-based configuration achieved a five-class accuracy point estimate of 71.5\%, compared with 69.9\% for the highest text-only zero-shot LLM result and 69.6\% for random forest. These small differences should be interpreted as descriptive point-estimate differences rather than evidence of statistical superiority. Vision combined with few-shot prompting also produced gains for selected models. Because each weather condition was represented by one fixed image, the results indicate that these specific images provided additional predictive signal for selected models; they do not establish generalization to unseen weather images or visual environments.

\textbf{Limitations and future research.}
The findings should be interpreted in light of several limitations. The sample consists of 92 students from one university and is not representative of the wider population. The analysis is based on stated choices rather than observed travel under actual weather conditions, and uncontrolled visual features within the generated images may have influenced both respondents and models. Each weather condition was represented by one fixed image, so weather condition and image identity were not separately identified. The relatively small and imbalanced dataset also limits detailed analysis of minority modes and traveler subgroups, while the results may vary across other LLM families and inference settings. The framework itself was not experimentally compared with a conventional research workflow; its contribution therefore concerns integration, traceability, and modularity rather than a measured reduction in survey cost or processing effort. In addition, the configuration comparisons were exploratory, and the highest reported accuracy may partly reflect the evaluation of multiple models and prompt settings on the same sample. Future research could evaluate the workflow with larger and more diverse samples, additional trip purposes and locations, and revealed-preference validation. Controlled image experiments could isolate the effects of weather, road condition, and surrounding environment. Further work could also examine prediction calibration, subgroup performance, and transferability across populations. The multi-agent workflow could support adaptive surveys and carefully validated synthetic respondents, but such outputs should be validated against human behavior before being used in transportation planning or policy analysis.

\section{Conclusions}

This study developed and evaluated a multi-agent workflow that integrates conversational survey design, AI-generated visual scenarios, structured data processing, conventional choice and machine-learning models, and LLM-based prediction within a unified travel-behavior research framework. The workflow was demonstrated using stated mode choices from 92 McGill student commuters across five image-augmented weather scenarios.

The empirical results support three main conclusions. First, the conversational survey produced behaviorally coherent responses across weather conditions, with adverse weather reducing cycling-related choices and increasing reliance on public transit. Second, the comparison of MNL, random forest, and LLM-based predictors highlighted their complementary roles. MNL provided interpretable behavioral relationships, random forest offered a strong trained predictive benchmark, and LLMs enabled flexible incorporation of habitual travel information, persona descriptions, examples, and visual context without requiring a separate model specification for each information type. Third, LLM performance depended strongly on prompt design and input configuration. Habitual travel information and expert-oriented framing generally improved prediction, while selected vision-capable models benefited from the inclusion of the same weather images shown to respondents. The best zero-shot LLM achieved a point estimate comparable to the trained machine-learning benchmark without task-specific fitting to the survey observations.

The main contribution of this study is therefore not the replacement of established travel-choice models, but the demonstration of an integrated and extensible workflow in which data collection, processing, behavioral analysis, predictive modeling, and multimodal reasoning can be coordinated through specialized agents. This framework provides a practical basis for incorporating heterogeneous behavioral information into travel-choice analysis while retaining conventional models for interpretation, benchmarking, and validation. More broadly, the findings indicate that LLMs may expand how traveler characteristics and contextual conditions are represented in transportation research, provided that their outputs are systematically evaluated against observed human behavior.


\section{Acknowledgments}

The authors used OpenAI ChatGPT and Claude (Anthropic) to assist with manuscript language editing and restructuring. All research-design decisions, data analyses, numerical results, references, interpretations, and final wording were reviewed by the authors. The LLM-based choice predictions evaluated in this study form part of the reported research methodology.

\section{Author Contributions}
All authors contributed to the study conception, analysis, and writing. All authors reviewed the results and approved the final version of the manuscript.

\section{Declaration of Conflicting Interests}

The authors declare that there is no conflict of interest regarding the publication of this paper.

\bibliographystyle{apalike}
\bibliography{references}

\end{document}